\documentclass[a4paper]{article}

\usepackage{INTERSPEECH2018}
\usepackage{amsmath}
\usepackage{paralist}

\title{Automatic Speech Recognition and Topic Identification for Almost-Zero-Resource Languages}
\name{  Matthew Wiesner$^{\dagger\ddagger}$, Chunxi Liu$^{\dagger\ddagger}$, Lucas Ondel$^{\S}$, Craig Harman$^{\dagger}$, Vimal Manohar$^{\dagger\ddagger}$, Jan Trmal$^{\dagger\ddagger}$\\ Zhongqiang Huang$^{\star}$, Najim Dehak$^{\dagger}$, Sanjeev Khudanpur$^{\dagger\ddagger}$ \thanks{This work was supported by DARPA LORELEI Grant N\b{o} HR0011-15-2-0024. The authors also thank Huda Khayrallah at the Johns Hopkins University for her help with the machine translation systems.} }
\address{
  $^\dagger$Center for Language and Speech Processing, The Johns Hopkins University, USA \\
$^\ddagger$Human Language Technology Center of Excellence, The Johns Hopkins University, USA \\
$^\S$Brno University of Technology, Czech Republic \qquad $^{\star}$Raytheon BBN Technologies, USA}
\email{\{wiesner,cliu77,londel1,charman,vmanoha1,yenda,khudanpur,ndehak3\}@jhu.edu, zhongqian.huang@raytheon.com}
\begin{document}

\maketitle
\begin{abstract}
Automatic speech recognition (ASR) systems often need to be developed for extremely low-resource languages to serve end-uses such as audio content categorization and search. While universal phone recognition is natural to consider when no transcribed speech is available to train an ASR system in a language, adapting universal phone models using very small amounts (minutes rather than hours) of transcribed speech also needs to be studied, particularly with state-of-the-art DNN-based acoustic models.  The DARPA LORELEI program provides a framework for such very-low-resource ASR studies, and provides an extrinsic metric for evaluating ASR performance in a humanitarian assistance, disaster relief setting. This paper presents our Kaldi-based systems for the program, which employ a universal phone modeling approach to ASR, and describes recipes for very rapid adaptation of this universal ASR system. The results we obtain significantly outperform results obtained by many competing approaches on the NIST LoReHLT 2017 Evaluation datasets.

\end{abstract}
\section{Introduction}
\label{sec:intro}

The goal of DARPA's Low Resource Languages for Emergent Incidents program (LORELEI) is the rapid development of human language technologies for low-resource languages, specifically in support of situational awareness for emergent missions such as humanitarian assistance, disaster relief, or response to an infectious disease outbreak \cite{strassel2016lorelei, malandrakis2017extracting}. The situational awareness gained from speech and text documents collected ``in the wild" is encoded in document descriptors called Situation Frames (SF). A SF consists of three elements that must be recognized, whenever present, in each speech document:
\begin{itemize}
\item {\bf Relevance} -- Produce a score of the document's relevance to the emergent incident,
\item {\bf Situation Type} -- Produce one or more of 11 predefined topics mentioned in the document,
\item {\bf Location} -- Extract any place names related to the incident mentioned in the document.
\end{itemize} 
The 11 topics were specified by the LORELEI program. 

%

The LORELEI SF detection task is characterized by extremely limited training resources. The only available resources for each evaluation language, called an Incident Language (IL) are:
\begin{enumerate}
	\item Monolingual text (only some of which is related to the incident)
    \item Untranscribed, unlabeled audio
    \item 10 hours of consultation with a native informant (NI).
	\item A small amount of IL-English parallel text 
\end{enumerate}
The NI is a native speaker of the IL with at least intermediate proficiency in English. System developers may ask the NI to perform any annotation tasks deemed necessary to build a system for extractng SFs from speech, e.g. transcribing speech or labeling documents with situation frames.

The lack of supervised training data in the IL demands the use of zero resource techniques, of cross-lingual knowledge transfer on many different levels, or of combinations thereof. To this end, we (i) developed an automatic speech recognition (ASR) system using {\it universal phone models}, (ii) explored {\it transfer of acoustic models} trained on closely related languages, and (iii) trained {\it language-independent classifiers} for situation types. These three approaches are the focus of this paper, and are applicable to other very-low-resource settings.

To obtain at least some labeled data in the IL|for adaptation of language universal systems|we asked the NI to read some IL text, transcribe some IL speech, and provide situation type  labels for some documents in the IL. To increase the NI's annotation efficiency, all NI tasks were conducted via a web browser-based user interface tailored to the specific LORELEI tasks.
We were able to obtain a few minutes (15-30) of transcribed IL speech and a few hundred (150-300) SF Type labels, which significantly improved performance. The read speech turned out to be useful for dianostic purposes during system development, but did not impact performance.



Other LORELEI project participants \cite{papadopoulos2017team} have used acoustic models trained on data collected during NI sessions and used an IL-to-English machine translation system and English-language SF-Type classifier. \cite{Littell2017} also train an English SF-Type classifier for this task, but translate the model's features to the IL, in which classification is then performed. As an alternative to such training of an ASR system from IL speech, we opted for a transfer learning paradigm and started with models trained on one or more higher-resource language(s). Other previous approaches \cite{loof2009cross,vu2011cross,mohan2012subspace,knill2014language} have explored cross-language ASR transfer assuming shared phonemic representations, generally using the \texttt{GlobalPhone} corpus \cite{schultz2002globalphone}, while \cite{ghoshal2013multilingual} examines multilingual training of a deep neural networks. Unlike these approaches, which had on the order of hours of target language speech, we are dealing with only minutes of adaptation speech.

In the remainder of the paper we describe the general system and its primary components. We describe the universal phone set ASR and language agnostic SF-Type classifier developed. Finally, we show results from the evaluation and analyze the extent to which adaptation of various components (using the data elicited from the NI) improves SF-Type task performance.

\section{General System}
\label{sec:system}
For the NIST LoReHLT 2017 evaluation the two ILs were Tigrinya (IL5) and Oromo (IL6). Both languages are spoken primarily in the Horn of Africa and are related to varying degrees to Amharic. For each IL two sets of audio data are provided: the development set called \texttt{set0 Speech}, and the evaluation set called \texttt{setE Speech}. The audio data consists of audio stories segmented into audio clips lasting no more than 2 minutes. For instance, the \texttt{set0 Speech} for IL5 consists of 83 audio stories segmented into 1323 audio clips; the \texttt{setE Speech} for IL5 consists of 116 audio stories segmented into 1095 audio segments. We refer to these audio clips as speech documents.

In our approach, we first convert the speech documents into sequences of tokens. The tokens can be words in the IL, or English translations of these words produced by a cascade of IL ASR and IL-to-English machine translation (MT). They can also be phone-like units discovered via acoustic unit discovery (AUD) \cite{ondel2016variational, liu2017empirical} or word-like units discovered via unsupervised term discovery (UTD) \cite{jansen2011efficient}.

\begin{figure}
\includegraphics[width=0.45\textwidth]{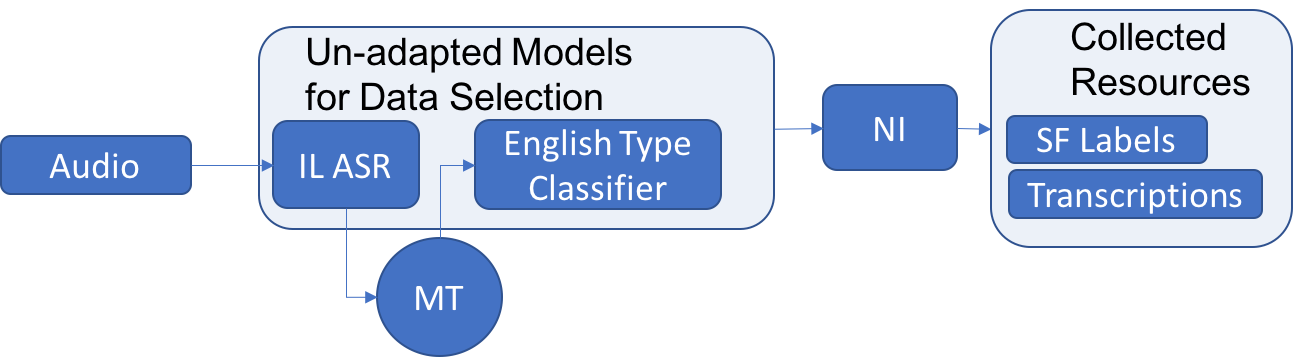}
\caption{Using the English SF-Type classifier to obtain adaptation/training data}\label{fig:NI_pipeline}

\end{figure}

We then select audio documents for transcription and/or SF-type annotation in order of their estimated informativeness.
After the NI has transcribed or annotated these documents, the transcriptions are used to adapt the ASR system and the SF-type annotations are added into the pool of training examples for the English SF-Type classifier. See Fig.~\ref{fig:NI_pipeline}.  Additionally, the labeled documents can be used to train three IL specific classifiers on the AUD, UTD, and IL word tokenizations of labeled \texttt{set0 Speech} audio documents respectively. In this way each tokenization scheme has a corresponding classifier capable of producing SF-Type scores for audio documents. 

\begin{figure}
\includegraphics[width=0.45\textwidth]{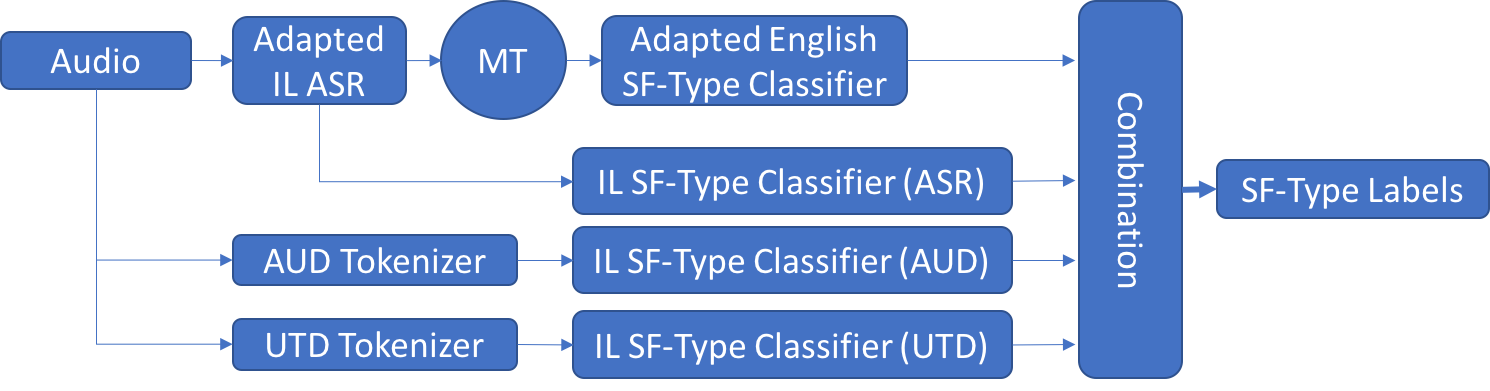}
\caption{SF-Type classification process}\label{fig:SF_classifier}

\end{figure}

Finally, for each of the four tokenizations of audio documents from \texttt{setE}, we use the corresponding SF-Type classifiers to produce SF-Type scores. Our final SF-Type scores are obtained as a weighted linear combination of the scores from the four different SF-Type classifiers. See the Fig.~\ref{fig:SF_classifier} 

\subsection{Data Selection}
The selection procedure described above 
relies heavily on English translations of the IL words. Each IL document can be classified using an English-language SF-Type classifier, trained in advance using only data from other languages. More precisely, we produce an SF-Type score for each document using the English-language SF-Type classifier. We then select the documents with the highest scores from each SF-Type to present to the NI for labeling (correcting) and/or transcription. We found that our data selection method outperforms random selection.


\section{Automatic Speech Recognition}
\label{sec:asr}

The two main obstacles to building ASR in the IL are training the acoustic models with little or no transcribed data and creating a suitable pronunciation lexicon. 

\vspace{-1em}\subsection{Acoustic Models}
\label{sec:acoustic_models}
The NI sessions are too short to collect enough data to train IL acoustic models from scratch. Hence, we depend on preexisting speech corpora to train acoustic models. All ASR systems were built using Kaldi \cite{povey2011kaldi}. We investigate acoustic model transfer from models trained on a single related language, as well as models trained on many unrelated languages.

In both methods of acoustic model transfer, some ad-hoc manual work may be required to 
map extra phones from one language to another. It is then possible to rebuild the ASR decoding graph by providing both an IL pronunciation lexicon and IL language model (LM). In both cases, a small amount of transcribed data can be used for subsequent acoustic model adaptation. 

\vspace{-1em}\subsubsection{Universal Phone Set ASR}
\label{sec:universal_phoneset_asr}
We refer to the transfer of acoustic models trained on many languages sharing a common phonemic representation as universal phone set ASR. Our approach is similar to \cite{knill2014language}. We use a selection of 10 BABEL languages for training, 7 of which were chosen as in \cite{knill2014language}, with 3 more chosen arbitrarily (Guarani, Mongolian, Dholuo). Diphthongs and triphtongs are split into their constituent phones to reduce the number, and enforce sharing, of phonemes. Also, as in \cite{knill2014language}, we standardize the representation of tone (tonal trajectory) across all training languages.
The final acoustic models are time-delay neural networks (TDNNs, \cite{peddinti2015time}) trained using the LF-MMI criterion (\cite{povey2016purely}).

\vspace{-1em}\subsubsection{Acoustic Model Adaptation}
We used a weights transfer approach for model adaptation from source to target language using transcribed data collected during the NI sessions. 
We used the same method that was used in \cite{manohar2017mgb3}.

\subsection{Pronunciation Lexicons and Language Models}
We bootstrapped the lexicon using a G2P trained on a seed lexicon derived from the provided resources.
For IL5 (Tigriyna) the seed was a dictionary of words with IPA pronunciations,
and for IL6 (Oromo) the seed  was an approximate grapheme-to-phoneme map. 

The vocabulary (word list) was generated from the provided monolingual text. We (re)normalized the text according to IL specific punctuation rules. Additional sources of words were the bilingual gazetteer, transcripts obtained during the NI sessions, and any provided dictionaries.
The LM was trained on the same text. LM hyper-parameters were chosen to minimize perplexity on a held-out set (small subset of the monolingual text not used for LM training).


\section{Situation Frame Type Classifiers}
\label{sec:sf classifiers}
We use two different approaches for Situation Frame classification. The first, based on IL tokenizations, requires SF-Type labels obtained during the NI sessions, but no IL MT. The second is a cross-lingual approach based on English tokenizations, requiring machine translation, but no IL SF-Type labels.

\subsection{IL Classifier}
After we tokenize the speech (see section \ref{sec:system}) we represent each speech document as a bag-of-words on unigram or $n$-gram occurrence counts of the tokens. Each vector is then scaled by the inverse document frequency (IDF) and normalized to $\mathcal{L}^2$ norm unit length.

For each SF type, a single classifier is trained as in \cite{liu2017topic}. Specifically, we use a set of 11 SVMs (Support Vector Machine classifiers), one for each type, trained on the bag-of-words features. We used stochastic gradient descent (SGD) based linear SVMs with hinge loss and $\mathcal{L}^2$ norm regularization~\cite{shalev2007pegasos, scikit-learn}. The SF-Type labels used for classifier training were obtained during the NI sessions. 

\subsection{English Classifier}
\label{sec:Eng_Classifier}
If no IL SF-type labels are available we can still leverage the
existing speech corpora of other development languages, which are annotated for SF-Type, in order to
train a universal SF-Type classifier. 
For each development language\footnote{Turkish (LDC2016E109), Arabic (LDC2016E123), Spanish (LDC2016E127), and Mandarin (LDC2016E108)}, we can construct an ASR system using existing ASR training data, transcribe the documents and translate the transcripts to English. 
After that, a single SF-Type classifier can be trained on the combined data.

In our system, we translate each word into its four most likely English
translations according to the probabilistic bilingual translation table
employed in the MT system that was developed for a separate LoReHLT MT
evaluation. The translation table is derived from the provided parallel
training data with words aligned automatically by the GIZA++~\cite{giza++}
and Berkeley aligner~\cite{berkeley_aligner}. In addition to using the
training data provided for the evaluation, native informants were also
consulted (independently under the MT effort) to produce hundreds of
parallel sentences and word translation pairs that are used in
training to increase the coverage of the MT system. 

We then produce bag-of-words features over English words.  If or when the SF-type labels of some IL documents become available, we can simply add these into the training data.

\section{Experiments}
\label{sec:eval}

Table \ref{table:NIResources} summarizes the resources collected during the NI sessions.
\begin{table}
\small
\caption{Overview of resources gathered during the NI sessions}
\centering
\begin{tabular}{c c c c}
\hline\hline
& Read & Transcribed & Labeled documents \\ [0.5ex]
\hline
IL5 & 20  mins & 27 mins & 159 \\
IL6 & 31  mins & 18  mins& 364 \\
\end{tabular}
\label{table:NIResources}
\end{table}
We use this data to adapt the systems described in sections~\ref{sec:asr} and ~\ref{sec:sf classifiers}. The labeled documents were used to train the IL SF-Type classifiers on UTD, AUD, and ASR tokenizations. We performed AUD as described in \cite{ondel2016variational}, but with two major modifications. First, the HMM model was embedded in a neural network generative model, known as Variational AutoEncoder (VAE) \cite{kingma2013auto}. Second, the model was initially trained supervisedly on a subset of the BABEL Amharic training data. For both incident languages, the model (VAE-HMM) was re-trained unsupervisedly. We performed both AUD and UTD on multilingual TDNN-based bottleneck features \cite{liu2017topic} of audio segments corresponding to speech. The segments were obtained from a DNN-based speech activity detection system that segmented audio into speech and silence. We also processed only speech segments when decoding the adapted IL5 ASR as this gave a slight improvement in performance.



For both IL5 and IL6 we treated Amharic as the related language and we trained a TDNN-LSTM system on the BABEL Amharic corpus. We generated triphone alignments as in \ref{sec:universal_phoneset_asr}. Our final IL5 system used the Amharic ASR, though we later found the adapted Universal model performed better. Our final IL6 system used the universal phone set ASR. Both systems were adapted using the collected transcribed speech. An adapted English SF-Type classifier for each language was trained by including all collected SF-type labels in the specific language. We used the read speech to evaluate the quality of both adapted and unadapted ASRs in both languages, as shown in table \ref{table:ASR}. Systems were evaluated on the \texttt{setE} \texttt{Speech} in two layers: the \emph{Relevance} layer (to separate the documents with at least 1 SF from non-relevant documents with zero SF present), and \emph{Type} layer (to detect all present SF types), using average precision (AP, equal to the area under the precision-recall curve). More evaluation metric details can be found in \cite{malandrakis2017extracting}.

\begin{table}[!htb]
\footnotesize
\caption{ASR Impact on SF-type Detection}
\centering
\begin{tabular}{c c c c}
\hline\hline
ASR & SF-Type & SF-Relevance & WER \\ [0.5ex]
\hline
IL5 Universal& 0.22 & 0.44 & 75.9 \\
IL5 Related& 0.26 & 0.46 & 68.5 \\
IL5 Adapt Related& 0.34 & 0.54 & 53.7\\
\textbf{IL5 Adapt Universal}& \textbf{0.35} & \textbf{0.54} & \textbf{51.6}\\ [0.3ex]
\hline\hline
IL6 Universal& 0.34 & 0.73 & 63.0 \\
IL6 Related& 0.35 & 0.74 & 47.9\\
IL6 Adapt Related& 0.37 & 0.77 & 44.4\\
\textbf{IL6 Adapt Universal}& \textbf{0.37}	& \textbf{0.77} & \textbf{39.8}\\ [1ex]
\end{tabular}
\label{table:ASR}
\end{table}

Table \ref{table:ILResults} shows the performance of our final submission systems. All ASR systems are adapted, and ASR+MT refers to the system using the English SF-Type classifier described in section \ref{sec:Eng_Classifier}.

\begin{table}[!htb]
\footnotesize
\caption{IL5 and IL6 Final Results}
\centering
\begin{tabular}{c|cc|cc}
\hline\hline
& \multicolumn{2}{c|}{IL5} & \multicolumn{2}{c}{IL6} \\
Method & \footnotesize{SF-Type} & \footnotesize{SF-Relevance} & \footnotesize{SF-Type} & \footnotesize{SF-Relevance} \\
 \hline
ASR+MT & 0.34 & 0.54 & 0.37 & 0.77 \\
ASR & 0.26 & 0.56 & 0.38 & 0.76 \\
AUD & 0.11 & 0.41 & 0.34 & 0.80 \\
UTD & 0.10 & 0.44 & 0.27 & 0.76 \\
\hline
\textbf{Combined} & \textbf{0.35} &  \textbf{0.58} & \textbf{0.41}  & \textbf{0.80}
\end{tabular}
\label{table:ILResults}
\end{table}

\subsection{ASR Adaptation}
Table \ref{table:ASR} compares the performance of the related-language and the universal phone set ASR before and after adaptation. ASR adaptation on the 15-30 min of collected transcribed speech improves SF-type classification modestly. Furthermore, WER seems to track SF-type  classification, which supports the utility of the SF-type task as an extrinsic measure of ASR performance. We also see that the universal phone set ASR has a similar WER to the adapted related language ASR when adapted on only 15-30 min of transcribed speech. 

While ASR adaptation resulted in large gains in IL5 (59\% SF-Type, 23\% SF-Relevance relative improvement), it helped only marginally in IL6 despite similar WER gains in both languages. Possible explanations are the smaller amount of IL6 adaptation data collected and/or MT quality (BLEU-4 ≈ 0.16 vs. BLEU-4 ≈ 0.09 for IL5/6 respectively).

\subsection{Classifier Adaptation}

The English SF-Type classifier was the best performing system (see row 1 of table~\ref{table:ILResults}).
For IL5, it was the best performing system by a wide margin, indicating that SF-Type labels derived from datasets from other languages can be extremely beneficial. We also examined how using the SF-Type labels from other languages affects performance. Table \ref{table:labels_ASR+MT}, shows how including various types of labels in training impacts performance.

\setlength{\tabcolsep}{0.10cm}
\begin{table}[!htb]
\small
\caption{IL SF-Type labels impact on SF-Type Classifiers. Adapted ASR, is the ASR used in the evaluation. MT is the IL-to-English MT described in \ref{sec:Eng_Classifier} using SF-Type labels ($\sim$ 3000) from other languages. Labels refers to IL specific labels collected from the NI. }
\vspace{-0.1cm}
\centering
\begin{tabular}{c|cc|cc}
\hline\hline
& \multicolumn{2}{c|}{IL5} & \multicolumn{2}{c}{IL6} \\
System & \footnotesize{Type} & \footnotesize{Rel} & \footnotesize{SF-Type} & \footnotesize{Rel} \\
\hline
Adapted ASR + MT + Labels & 0.35 & 0.54 & 0.37 & 0.77 \\
Adapted ASR + MT + No Labels & 0.26 & 0.46 & 0.19 & 0.73 \\
Adapted ASR + Labels & 0.26 & 0.56 & 0.38 & 0.77
\end{tabular}
\label{table:labels_ASR+MT}
\end{table}

\begin{figure}[!htb]
 \centering
\includegraphics[width=0.4\textwidth]{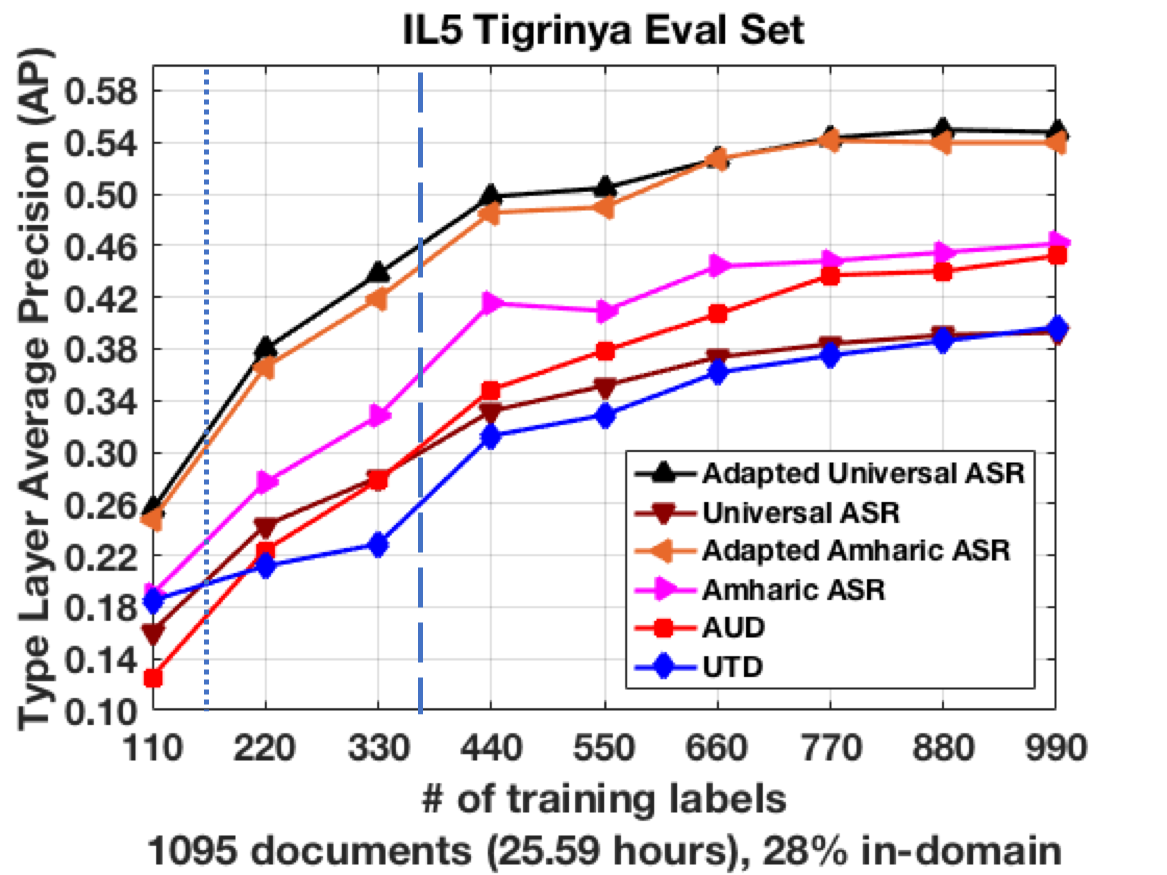}
\includegraphics[width=0.4\textwidth]{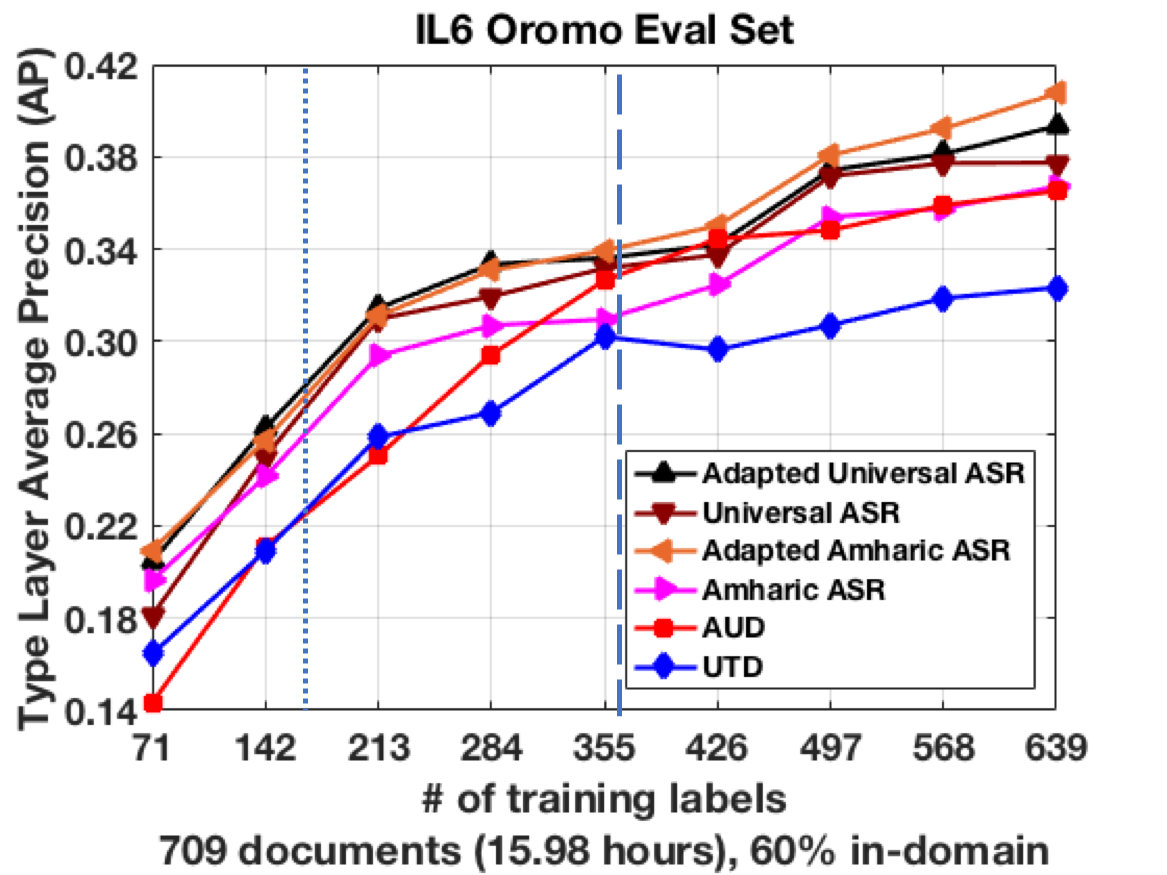}
\caption{IL5,6 SF-Type Classifier performance as a function of the number of SF-Type labels in training. The vertical dotted line shows the number of SF-Type labels collected from the NI in Tigrinya (IL5). The vertical dashed line shows the number of SF-Type labels collected in Oromo (IL6). Since the SF-Type labels used are from \texttt{setE} \texttt{Speech}, there is a small discrepancy in type and relevance scores compared to the evaluation results.}
\label{fig:Classifier_adaptation}
\end{figure}
\vspace{-0.1cm}

We note that using the English SF-Type classifier trained only on the combined set of 3000 SF-Type labels from the development languages (row 2 of table \ref{table:labels_ASR+MT}) yields similar performance in IL5 as training an IL SF-Type classifier (row 3 of table \ref{table:labels_ASR+MT}) on only 159 IL specific SF-Type labels.
While the English SF-Type classifier performed significantly worse on IL6 results (row 2 of table \ref{table:labels_ASR+MT}), we believe that the English SF-Type classifier trained on labels from other languages can match the performance of an IL-specific SF-Type classifier. 
However, adding the IL specific SF-Type labels to the English SF-Type classifier training data always improves performance (rows 1,3 of table \ref{table:labels_ASR+MT}). 

To demonstrate the value of IL specific SF-Type labels we performed the following experiment on the \texttt{setE} \texttt{Speech} ground truth SF-Type labels of both IL5 and IL6. For each language, and each of 6 tokenizations (see Fig. \ref{fig:Classifier_adaptation}) we trained IL specific SF-Type classifiers, varying the number of SF-Type labels used in training. We split the \texttt{setE} \texttt{Speech} of each language into 10 folds and measured the performance, by 10-fold cross validation, of each SF-Type classifier trained on between 1 and 9 folds worth of labels. Figure \ref{fig:Classifier_adaptation} shows the results of this experiment.
 

We see from figure \ref{fig:Classifier_adaptation} that IL5 and IL6 SF-Type classifiers trained on the same number of IL SF-Type labels perform similarly for AUD, UTD and unadapted ASR tokenizations; the IL6 AUD and UTD systems likely outperformed the corresponding IL5 systems because we collected more IL6 specific SF-Type labels. Collecting more IL specific SF-Type labels always helps performance. We also see in IL5 that adding 159 SF-Type labels to training ($\sim$ 2h NI time) is comparable to ASR-adaptation on 27 min of transcribed speech ($\sim$ 6h NI time).

\section{Conclusions}
This paper presents an SF-Type classification system of speech documents used in the LoReHLT 2017 evaluation. The system combines universal acoustic modeling, IL-to-English machine translation (MT) and an English-language topic classifier. This combination requires no transcribed speech in the evaluation language, leading to near language-agnostic operation. We demonstrated that adaptation on a small amount of transcribed speech yields modest improvement in SF-type classification. However, with enough IL specific SF-Type labels, an MT-free system can achieve the same performance. 

Finally we must consider that the intrinsic value of ASR-based systems lies in the semantically meaningful tokenization they produce. Using ASR-based systems opens up a promising venue of research directed towards detecting names of people and places in speech. This can be formulated as a keyword search task using word-based search \cite{trmal2014keyword, trmal2017kaldi}, phonetic-based search, or a fusion of the two \cite{liu2014low}.




\bibliographystyle{IEEEtran}

\bibliography{mybib}

\end{document}